\documentclass[11pt]{article}

\usepackage[]{acl}

\usepackage{times}
\usepackage{latexsym}

\usepackage[T1]{fontenc}

\usepackage[utf8]{inputenc}

\usepackage{microtype}

\usepackage[utf8]{inputenc} 
\usepackage[T1]{fontenc}    
\usepackage{hyperref}       
\usepackage{url}            
\usepackage{booktabs}       
\usepackage{amsfonts}       
\usepackage{nicefrac}       
\usepackage{microtype}      
\usepackage{xcolor}         
\usepackage{multirow}
\usepackage{CJKutf8}        
\usepackage[pdftex]{graphicx}
\nointerlineskip 
\usepackage{subfigure}
\usepackage{makecell}
\usepackage{amsmath}
\usepackage{makecell}
\usepackage{pdfpages}
\usepackage{indentfirst}
\usepackage{enumerate}
\usepackage{multirow}
\usepackage{graphicx}

\title{Mining Clues from Incomplete Utterance:\\ A Query-enhanced Network for Incomplete Utterance Rewriting}

\author{Shuzheng Si$^{1, 2}$\footnotemark[1], Shuang Zeng$^{1,2}$\footnotemark[1], Baobao Chang$^{1}$\footnotemark[2] \\
$^1$Key Laboratory of Computational Linguistics, Peking University, MOE, China \\ 
$^2$School of Software and Microelectronics, Peking University, China \\
\texttt{sishuzheng@stu.pku.edu.cn \{zengs,chbb\}@pku.edu.cn}
}

\begin{document}

\maketitle


\definecolor{highlight}{rgb}{0,0,0}

\begin{abstract}
Incomplete utterance rewriting has recently raised wide attention.
However, previous works do not consider the semantic structural information between incomplete utterance and rewritten utterance or model the semantic structure implicitly and insufficiently.
To address this problem, we propose a \textbf{QUE}ry-\textbf{E}nhanced \textbf{N}etwork (\textbf{QUEEN}).
Firstly, our proposed query template explicitly brings guided semantic structural knowledge between the incomplete utterance and the rewritten utterance making model perceive where to refer back to or recover omitted tokens.
Then, we adopt a fast and effective edit operation scoring network to model the relation between two tokens.
Benefiting from proposed query template and the well-designed edit operation scoring network, QUEEN achieves state-of-the-art performance on several public datasets. 
\end{abstract}
\renewcommand{\thefootnote}{\fnsymbol{footnote}}
\footnotetext{$^*$Equal contribution.}
\footnotetext{$^\dag$Corresponding author.}
\renewcommand{\thefootnote}{\arabic{footnote}}

\section{Introduction}

\begin{CJK}{UTF8}{gbsn}
{\color{highlight}

\noindent
Multi-turn dialogue modeling, a classic research topic in the field of human-machine interaction, serves as an important application area for pragmatics \citep{leech2003pragmatics,li2023pace,si2023spokenwoz} and Turing Test. 
The major challenge in this task is that interlocutors tend to use incomplete utterances for brevity, such as referring back to (i.e., coreference) or omitting (i.e., ellipsis) entities or concepts that appear in dialogue history. As shown in Table \ref{tab_1}, the incomplete utterance $u_3$ refers to ``Smith'' (``史密斯'') from $u_1$ and $u_2$ using a pronoun ``He'' (``他'') and omits ``the type of cuisine'' (``菜肴的类型'') from $u_2$. 
This may cause referential ambiguity and semantic incompleteness problems if we only read this single utterance $u_3$, which is a common case of downstream applications like retrieval-based dialogue systems \citep{Retrieval-based-Dialogue-Systems}.
Moreover, previous studies \citep{su-etal-2019-improving, pan-etal-2019-improving} also find that coreference and ellipsis exist in more than 70\% of the utterances, especially in pro-drop languages like Chinese.
These phenomena make it imperative to effectively model dialogue in incomplete utterance scenarios.

\begin{table}
\small
\setlength{\tabcolsep}{1.0mm}{
\renewcommand{\arraystretch}{1.1}
\begin{tabular}{cc}
{Turns} & {Utterances with Translation}\\
\hline
\multirow{2}{*}{$u_1$} & {\textcolor{blue}{Smith} needs to find an expensive restaurant nearby.} \\
& {\textcolor{blue}{史密斯}需要在附近找一家昂贵的餐馆。} \\
\hline
\multirow{2}{*}{$u_2$} & {Does \textcolor{blue}{Smith} care
\textcolor{red}{the type of cuisine}?} \\
& {\textcolor{blue}{史密斯}关心\textcolor{red}{菜肴的类型}吗？} \\
\hline
\multirow{2}{*}{$u_3$} & {No, \textcolor{blue}{he} does not care.} \\
& {不，\textcolor{blue}{他}不关心。} \\
\hline
\multirow{2}{*}{$u^{'}_3$} & {No, \textcolor{blue}{Smith} does not care about \textcolor{red}{the type of cuisine}.} \\
& {不，\textcolor{blue}{史密斯}不关心\textcolor{red}{菜肴的类型}。} \\
\hline
\end{tabular}}
\caption{An example in multi-turn dialogue including dialogue utterance history $u_1$ and $u_2$, incomplete utterance $u_3$ and rewritten utterance $u_3'$.}
\label{tab_1}
\end{table}

To cope with this problem, previous works \citep{kumar-joshi-2016-non,elgohary-etal-2019-unpack,su-etal-2019-improving} propose the \textbf{I}ncomplete \textbf{U}tterance \textbf{R}ewriting (\textbf{IUR}) task. It aims to rewrite an incomplete utterance into a semantically equivalent but self-contained utterance by mining semantic clues from the dialogue history. Then the generated utterance can be understood without reading dialogue history.
For example, in Table \ref{tab_1}, after recovering the referred and omitted information from $u_3$ into $u_3'$, we could better understand this utterance comprehensively than before. 

Early works use coreference resolution methods \citep{clark-manning-2016-improving} to identify the entity that a pronoun refers to. 
However, they ignore the more common cases of ellipsis.
So the text generation-based methods \citep{su-etal-2019-improving,pan-etal-2019-improving} are introduced to generate the rewritten sequence from the incomplete sequence by jointly considering coreference and ellipsis problems.
Though effective, generation models neglect a key trait of the IUR task, where the main semantic structure of a rewritten utterance is usually similar to the original incomplete utterance. So the inherent structure-unawareness and uncontrollable feature of generation-based models impede their performances. 
For semantic structure-aware methods, \citet{qian2020incomplete} utilize an edit operation matrix (e.g., substitution, insertion operations) to convert an incomplete utterance into a complete one. They formulate this task as a semantic segmentation problem with a CNN-based model \citep{ronneberger2015u} on the matrix to capture the semantic structural relations between words implicitly. 
\citet{xu-etal-2020-semantic} attempt to add additional semantic information to language models \citep{DBLP:conf/naacl/DevlinCLT19} by annotating semantic role information but it is time-consuming and costly. 
\citet{DBLP:conf/aaai/HuangLZ021} propose a semi-autoregressive generator using a tagger to model the some considerable overlapping regions between the incomplete utterance and rewritten utterance, yet only implicitly learn the difference between them.
Although these methods maintain some similarities between the incomplete utterance and the rewritten utterance (i.e., the overlap between them), it is difficult for these methods to explicitly model the semantic structure, especially the difference between the two utterances, ignoring the information in the incomplete utterance, such as which tokens are more likely to be replaced and which positions are more likely to require the insertion of new tokens.
Therefore, there are still limitations of existing methods for IUR task, especially in jointly considering coreference and ellipsis cases and better utilizing semantic structural information. 

This paper proposes a simple yet effective \textbf{QUE}ry-\textbf{E}nhanced \textbf{N}etwork (\textbf{QUEEN}) to solve the IUR task. QUEEN jointly considers coreference and ellipsis problems that frequently happen in multi-turn utterances. Specifically, we propose a straightforward query template featuring two linguistic properties and concatenate this query with 
utterances as input text. This query explicitly brings semantic structural guided information shared between the incomplete and the rewritten utterances, i.e., making model perceive where to refer back to or recover omitted tokens. %
We regard the rewritten utterance as the output from a series of edit operations on the incomplete utterance by constructing a token-pair edit operation matrix, which attempts to model the the overlap between the incomplete utterance between the rewritten utterance. 
Different from \citet{qian2020incomplete},
we adopt a well-designed edit operation scoring network on the matrix to perform incomplete utterance rewriting, which is faster and more effective. 
QUEEN brings semantic structural information from linguistics into the model more explicitly and avoids unnecessary overheads of labeled data from other tasks. Experiments on several IUR benchmarks show that QUEEN outperforms previous state-of-the-art methods.
Extensive ablation studies also confirm that the proposed query template
makes key contributions to the improvements of QUEEN.

\begin{figure*}[]
\centering 
\includegraphics[width=0.9\linewidth]{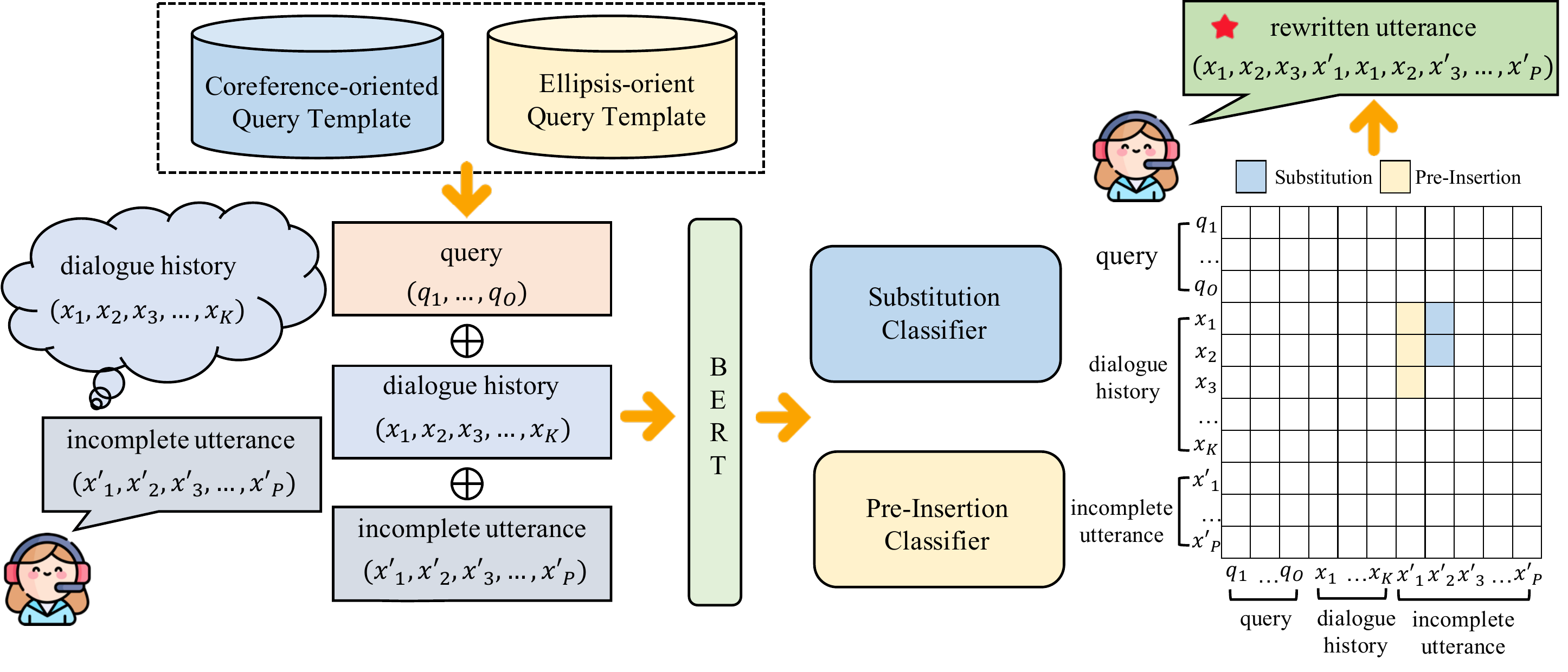} 
\caption{General architecture of QUEEN.
} 
\label{Fig_new} 
\end{figure*}

}
\end{CJK}
\section{Methodology}

\begin{CJK}{UTF8}{gbsn}
{\color{highlight}

\noindent
\paragraph{Overview} Our QUEEN mainly consists of two modules: query template construction module (Sec.~\ref{sec:query template}) and edit operation scoring network module (Sec.~\ref{sec:model architecture}). 
From two linguistic perspectives, the former module aims to generate a query template for each incomplete utterance, i.e., coreference-ellipsis-oriented query template, to cope with coreference and ellipsis problems. 
This query template explicitly hints the model where to refer back and recover omitted tokens.
The latter module tries to capture the semantic structural relations between tokens by constructing an edit operation matrix.
As shown in Figure \ref{Fig_new}, our goal is to learn a model to generate correct edit operations on this matrix and compute edit operation scores between token pairs so as to convert the incomplete utterance into the complete one.

\subsection{Query Template Construction Module} 
\label{sec:query template}
\noindent
By observing incomplete and rewritten utterance pairs in existing datasets, we find that pronouns and referential noun phrases in the incomplete utterance often need to be substituted by text spans in dialogue history.
And ellipsis often occurs in some specific positions of incomplete utterance, conforming to a certain syntactic structure. In this module, we expect to encode these linguistic prior knowledge into the input of QUEEN. The query template is constructed as follows:

\noindent
\paragraph{Coreference-oriented Query Template}
In order to make QUEEN perceive the positions of coreference that need to be substituted by text spans from dialogue history, we use a special token \texttt{[COREF]} to replace pronouns and referential noun phrases in the incomplete utterance so as to get our coreference-oriented query template. For example, the coreference-oriented query template of the incomplete utterance 
``No, he does not care'' (``不,他不关心'') is ``No, \texttt{[COREF]} does not care'' (``不,\texttt{[COREF]}不关心'') .
To get the target complete utterance, this query explicitly tells the model we should replace the ``He'' (``他'') with text spans (such as 'Smith'(``史密斯'') ) from dialogue history, rather than replacing other words. 
Here, we find all pronouns that required to be replaced using a predefined pronoun collection.

\noindent
\paragraph{Ellipsis-oriented Query Template} To make QUEEN perceive the positions of ellipsis that need to be inserted by text spans from dialogue history, we define a special token \texttt{[ELLIP]} and put it in a linguistically right place of the incomplete utterance. 
Since a self-contained utterance usually contains a complete S-V-O (Subject-Verb-Object) structure, if an incomplete utterance lack any of these key elements, we could assume there is a case of ellipsis in its corresponding text position. So we perform dependency parsing on the incomplete utterance to get the structure of the incomplete utterance.
For example, the parsing result of the incomplete utterance ``No, he does not care'' (``不，他不关心'') is an S-V structure and lack object element, thus we put \texttt{[ELLIP]} at the end of the sentence to get the ellipsis-oriented query template as ``No, he does not care \texttt{[ELLIP]}'' (``不,他不关心\texttt{[ELLIP]}''). 

Then we fuse these two query templates into the final coreference-ellipsis-oriented query template. For incomplete utterance ``No, he does not care'' (``不,他不关心''), we get ``No, \texttt{[COREF]} does not care \texttt{[ELLIP]}' (``不,\texttt{[COREF]} 不关心 \texttt{[ELLIP]}'') as our final query template. Under supervised setting, the models will perceive the positions to refer back and recover omitted tokens for this utterance. 
For a multi-turn dialogue $d=(u_1,...,u_{N-1},u_N)$ containing $N$ utterances where $u_1\sim u_{N-1}$ are dialogue history and the last utterance $u_N$ needs to be rewritten, 
we could get the dialogue history text $s = (w^{1}_{1},...,w^{n}_{i},...,w^{N}_{L_{N}})$ where $w^{n}_{i}$ is the $i$-th token in the $n$-th utterance and $L_{n}$ is the length of $n$-th utterance. We then concatenate our coreference-ellipsis-oriented query template with the dialogue history text to get our final input text $s' = (w^{q}_{1},...,w^{q}_{k},...,w^{q}_{M}, w^{1}_{1},...,w^{n}_{i},...,w^{N}_{L_{N}})$ where $w^{q}_{k}$ is the $k$-th token of the query template and $M$ is the length of query template.
}
\end{CJK}


\subsection{Edit Operation Scoring Network Module}
\label{sec:model architecture}
\noindent
Since pre-trained language models have been proven to be strongly effective on several natural language processing tasks, we employ BERT \citep{DBLP:conf/naacl/DevlinCLT19} to encode our input text to get the contextualized hidden representation
$H=(h^{q}_{1},...,h^{q}_{k},...,h^{q}_{M}, h^{1}_{1},...,h^{n}_{i},...,h^{N}_{L_{N}})$.
Our model attempts to predict whether there is an edit operation between each token pair. To this end, we define an operation scoring function as follows. Since the order of utterance is also important for dialogue, we further use RoPE \citep{su2021roformer} to provide relative position information :
\begin{align}
q^{\alpha}_{i} &= W^{\alpha}h_i + b^{\alpha}\\
k^{\alpha}_{j} &= W^{\alpha}h_j + b^{\alpha}\\
{s}^{\alpha}_{ij} &= (R_{i}q^{\alpha}_{i})^T(R_{j}k^{\alpha}_{j})
\end{align}
where $\alpha$ is edit operation type including \textit{Substitution} and \textit{Pre-Insertion}. For different operations, we use different trainable parameters $W^{\alpha}$ and $b^{\alpha}$. 
$R$ is a transformation matrix from RoPE to inject position information and ${s}^{\alpha}_{ij}$ is the score for $\alpha$-th edit operation from $i$-th token in dialogue history to $j$-th token in incomplete utterance.
 
During decoding for $\alpha$-th operation, edit operation label $\mathcal{Y}^{\alpha}_{ij}$ satisfies:
\begin{align}
\mathcal{Y}^{\alpha}_{ij}=\left\{
\begin{array}{rcl}
1 & {{s}^{\alpha}_{ij} >= \theta}\\
0 & {{s}^{\alpha}_{ij} < \theta}
\label{con:decoding}
\end{array} \right.
\end{align}
where $\theta$ is a hyperparameter. Once $\mathcal{Y}^{\alpha}_{ij}$ equals to 1, the edit operation $\alpha$ should be performed between token $i$ and token $j$.

Since the label distribution of edit operation is very unbalanced (most elements are zeros), we employ Circle Loss \citep{sun2020circle, DBLP:journals/corr/abs-2208-03054} to mitigate this problem:

\begin{footnotesize}
\begin{align}
log(1+\sum_{(i,j)\in \Omega_{pos}}e^{-s^{\alpha}_{i,j}})+log(1+\sum_{(i,j)\in \Omega_{neg}}e^{s^{\alpha}_{i,j}})
\end{align}
\end{footnotesize}

\noindent
where $\Omega_{pos}$ is the positive sample set for edit operation $\alpha$, and $\Omega_{pos}$ is the negative sample set.

\begin{table}
\small
\centering
\setlength{\tabcolsep}{2.1mm}{
\renewcommand{\arraystretch}{1.1}
\begin{tabular}{lccccc}
\hline
\textbf{Model} &{EM} & {$B_2$} & {$B_4$} & {$R_2$} & {$R_L$}\\
\hline
{T-Gen$^\dag$} & {35.4} & {72.7} & {62.5} & {74.5} & {82.9}\\
{L-Ptr-$\lambda ^\dag$}  & {42.3} & {82.9} & {73.8} & {81.1} & {84.1}\\
{L-Gen$^\dag$}  & {47.3} & {81.2} & {73.6} & {80.9} & {86.3}\\
{L-Ptr-Gen$^\dag$}   & {50.5} & {82.9} & {75.4} & {83.8} & {87.8}\\
{T-Ptr-Gen$^\dag$}  & {53.1} & {84.4} & {77.6} & {85.0} & {89.1}\\
{T-Ptr-$\lambda ^\dag$}  & {52.6} & {85.6} & {78.1} & {85.0} & {89.0}\\
{T-Ptr-$\lambda$+BERT$^\dag$}  & {57.5} & {86.5} & {79.9} & {86.9} & {90.5}\\
{CSRL} & {60.5} & {86.8} & {77.8} & {85.9} & {90.5}\\
{RUN$^\dag$} & {66.4} & {91.4} & {86.2} & {90.4} & {93.5}\\
{\textbf{QUEEN}} & \textbf{70.1} & \textbf{92.1} & \textbf{86.9} & \textbf{90.9} & \textbf{94.6}\\
\hline
\end{tabular}
}
\caption{{Experimental results on REWRITE. $\dag$: Results from \citet{qian2020incomplete}.}}
\label{tab_3} 
\end{table}

\begin{table}
\small
\centering
\setlength{\tabcolsep}{2.8mm}{
\renewcommand{\arraystretch}{1.1}
\begin{tabular}{lccccc}
\hline
\textbf{Model} & {EM} & {$B_1$} & {$B_2$} & {$R_1$} & {$R_2$}\\
\hline
{Syntactic$^\dag$} &{-} & {84.1} & {81.2} & {89.3} & {80.6}\\
{L-Gen$^\dag$}  & {-} & {84.9} & {81.7} & {88.8} & {80.3}\\
{L-Ptr-Gen$^\dag$}  & {-} & {84.7} & {81.7} & {89.0} & {80.9}\\
{BERT$^\ddag$}  &{-} & {85.2} & {82.5} & {89.5} & {80.9}\\
{PAC$^\dag$} &{-} & {89.9} & {86.3} & {91.6} & {82.8}\\
{CSRL$^\ddag$} &{-} & {85.8} & {82.9} & {89.6} & {83.1}\\
{SARG} &{-} & {92.2} & {89.6} & {92.1} & {86.0}\\
{RUN$^\dag$} &{49.3} & {92.3} & {89.6} & {92.4} & {85.1}\\
{\textbf{QUEEN}} & \textbf{53.5} & \textbf{92.4} & \textbf{89.8} & \textbf{92.5} & \textbf{86.3}\\
\hline
\end{tabular}}
\caption{Experimental results on Restoration-200K.  Additionally, we reproduce from the released code to get EM of RUN. $\dag$: Results from \citet{qian2020incomplete}. $\ddag$: Results from \citet{xu-etal-2020-semantic}.}
\label{tab_4} 
\end{table}

\begin{table}
\small
\centering
\setlength{\tabcolsep}{5.6mm}{
\renewcommand{\arraystretch}{1.1}
\begin{tabular}{lcc}
\hline
\textbf{Model} &{EM} & {$B_4$}\\
\hline
{Ellipsis Recovery$^\dag$}  & {50.4} & {74.1}\\
{GECOR 1$^\dag$}  & {68.5} & {83.9}\\
{GECOR 2$^\dag$}  & {66.2} & {83.0}\\
{RUN$^\dag$}  & {69.2} & {85.6}\\
{\textbf{QUEEN}} & \textbf{71.6} & \textbf{86.3}\\
\hline
\end{tabular}}
\caption{{Experimental results on TASK. $\dag$: Results and evaluation metrics from \citet{qian2020incomplete}.}}
\label{tab_TASK} 
\end{table}

\begin{table}
\small
\centering
\setlength{\tabcolsep}{2mm}{
\renewcommand{\arraystretch}{1.1}
\begin{tabular}{lcccccc}
\hline
\textbf{Model} &{$B_1$} &{$B_2$} &{$B_4$} & {$R_1$}& {$R_2$}& {$R_L$}\\
\hline
{Copy$^\dag$}  & {52.4} & {46.7}  & {37.8} & {72.7}  & {54.9} & {68.5}\\
{Pronoun Sub$^\dag$}  & {60.4} & {55.3}  & {47.4} & {73.1}  & {63.7} & {73.9}\\
{L-Ptr-Gen$^\dag$}  & {67.2} & {60.3}  & {50.2} & {78.9}  & {62.9} & {74.9}\\
{RUN$^\dag$}  & {70.5} & {61.2}  & {49.1} & {79.1}  & {61.2} & {74.7}\\
{\textbf{QUEEN}} & \textbf{72.4} & \textbf{65.2}  & \textbf{54.4} & \textbf{82.5}  & \textbf{68.1} & \textbf{81.8}\\
\hline
\end{tabular}}
\caption{{Experimental results on CANARD. $\dag$: Results and evaluation metrics from \citet{qian2020incomplete}.}}
\label{tab_CANARD} 
\end{table}

We tune Circle Loss the same as \citet{DBLP:conf/ijcai/ZhangCXDTCHSC21} and \citet{DBLP:journals/corr/abs-2208-03054}.
We refer readers to their paper for more details.

\section{Experiments}

\begin{CJK}{UTF8}{gbsn}
\subsection{Experimental Setup}
\noindent
\paragraph{Datasets} We evaluate our model on four IUR benchmarks from different domains and languages: REWRITE (Chinese, \citealp{su-etal-2019-improving}), Restoration-200K (Chinese, \citealp{pan-etal-2019-improving}), TASK (English, \citealp{quan2019gecor}), CANARD (English, \citealp{elgohary-etal-2019-unpack}). 
Some statistics are shown in Table \ref{tab_7}. REWRITE and Restoration-200K are constructed from Chinese Open-Domain Dialogue. TASK is from English Task-oriented Dialogue. CANARD is constructed from English Context Question Answering. We follow the same data split as their original paper.

\noindent
\paragraph{Evaluation} We use BLEU \citep{papineni-etal-2002-bleu}, ROUGE \citep{lin2004rouge} and the exact match (EM score) as our evaluation metrics.

\begin{table*}
\centering
\setlength{\tabcolsep}{4.mm}{
\renewcommand{\arraystretch}{1.1}
\begin{tabular}{lcccccccc}
\hline
\textbf{Variant} & {EM}&{$B_1$}&{$B_2$}&{$B_3$}&{$B_4$}&{$R_1$}&{$R_2$}&{$R_L$}\\
\hline
{\textbf{QUEEN}} & {\textbf{70.1}} & {\textbf{94.7}} & {\textbf{92.1}} & {\textbf{89.5}} & {\textbf{86.9}}& {\textbf{96.0}}& {\textbf{90.9}}& {\textbf{94.6}}\\
\hline
{   -w/o query} & {67.4} & {92.9} & {90.5} & {88.1} & {85.7}& {95.2}& {90.1}& {94.0}\\
{   -w/o CQT} &{67.9} & {93.5} & {91.0} & {88.5} & {86.0} & {95.5}& {90.4}& {94.2}\\
{   -w/o EQT} &{67.4} & {93.5} & {91.1} & {88.5} & {85.9} & {95.4}& {90.3}& {94.3}\\
\hline
\end{tabular}}
\centering
\caption{
The ablation results on the development set of REWRITE.
“w/o query” means that we do not append a designed query before encoding that semantic information into our model.
“w/o CQT” means that we only perform Ellipsis-oriented Query Template. 
“w/o EQT” means that we only perform Coreference-oriented Query Template and use the incomplete utterance as query if match fails.
}
\label{tab_ablation} 
\end{table*}

\noindent
\paragraph{Baseline Models} We compare our model with a large number of baselines and SOTA models. (i) \textit{Baselines and Generation models} include L-Gen \citep{bahdanau2014neural}, the hybrid pointer generator network (L-Ptr-Gen) \citep{DBLP:conf/acl/SeeLM17}, the basic transformer model (T-Gen) \citep{DBLP:conf/nips/VaswaniSPUJGKP17} and the transformer-based pointer generator (T-Ptr-Gen) \citep{DBLP:conf/acl/SeeLM17}, Syntactic \citep{kumar2016non}, PAC \citep{pan-etal-2019-improving}, L-Ptr-$\lambda$ and T-Ptr-$\lambda$ \citep{su-etal-2019-improving}, GECOR \citep{quan2019gecor}. 
Above methods need to generate rewritten utterances from scratch, neglecting the semantic structure between a rewritten utterance and the original incomplete one.
(ii) \textit{Structure-aware models} include CSRL \citep{xu-etal-2020-semantic}, RUN \cite{qian2020incomplete}, SARG \cite{DBLP:conf/aaai/HuangLZ021}. 

\noindent
\paragraph{Hyper-parameters} We implement our model on top of a BERT-base model \citep{DBLP:conf/naacl/DevlinCLT19}. We initialize QUEEN with bert-base-uncased for English and bert-base-chinese for Chinese. We use Adam \citep{DBLP:journals/corr/KingmaB14} with learning rate 1e-5. The batch size is set to 16 for REWRITE and TASK, 12 for Restoration-200K, 4 for on CANARD. Meanwhile, $\theta$ in Equation \ref{con:decoding} is set to 0.1 for REWRITE and TASK, 0.05 for Restoration-200K and CANARD.


\begin{table}
\setlength{\tabcolsep}{0.3mm}{
\small
\renewcommand{\arraystretch}{1.1}
\begin{tabular}{lcccc}
\hline
{} & {Restoration-200K} & {REWRITE} &{TASK} &{CANARD}\\
\hline
{Language} &{Chinese} &{Chinese} &{English} &{English}\\
{His Avglen} & {25.8} & {17.7} & {52.6} &{85.4}\\
{Inc Avglen} & {8.6} & {6.5} &{9.4} &{7.5}\\
{Rew Avglen} & {12.4} & {10.5} &{11.3} &{11.6}\\
\hline
\end{tabular}}
\caption{Statistics of different datasets. 'Avglen' for average length, 'His' for historical utterance, 'Inc' for incomplete utterance, and 'Rew' for rewritten utterance.}
\label{tab_7}
\end{table}

\subsection{Experimental Details}
\label{sec:appendix}

\paragraph{Constructing Supervision} The expected supervision for our model is the edit operation matrix, but existing datasets only contain rewritten utterances. So we adopt Longest Common Subsequence (LCS) and 'Distant Supervision'
\cite{qian2020incomplete} to get correct supervision, which contains edit operations \textit{Substitute} and \textit{Pre-Insert}.

\noindent
\paragraph{Coreference-oriented Query} During the training stage, We use the ground truth of pronouns and referential noun phrases to construct the coreference-oriented query.
During the inference, we use the constructed pronoun collection to construct the coreference-oriented query, which contains pronouns and referential noun phrases from training data and common pronouns.

\noindent
\paragraph{Ellipsis-oriented Query Construction} If the parsing result of the incomplete utterance is an S-V (Subject-Verb) structure and lacks subject element, we insert an \texttt{[ELLIP]} at the end of the incomplete utterance as the query. When there is not the S-V structure after parsing, we insert an \texttt{[ELLIP]} at the beginning of the incomplete utterance as the query. In other cases, we insert \texttt{[ELLIP]} at both the beginning and end of the incomplete utterance as the query. We use spaCy\footnote{https://spacy.io/} for English and LTP \citep{che2020n} for Chinese to get the result of parsing.

\noindent
\textbf{Extra Findings} \quad During the experiment, we find two interesting points: (i) As \texttt{[COREF]} and \texttt{[ELLIP]} are sparse respectively, we use a unified token \texttt{[UNK]} to replace \texttt{[COREF]} and \texttt{[ELLIP]} in the query to relieve the sparsity. (ii) In most cases, if there is the referring back in the utterance, there is generally no ellipsis in the utterance. Redundant \texttt{[ELLIP]} tokens can't bring correct guided information in this case. Therefore, once we construct Coreference-oriented Query Template successfully, we will not try to construct the Ellipsis-oriented  Query Template. Our experimental results are improved by the above two tricks.

\subsection{Results and Analysis}
\noindent
\paragraph{Main Results} We report the experiment results in Table \ref{tab_3}, Table \ref{tab_4}, Table\ref{tab_TASK} and Table \ref{tab_CANARD}.
On all datasets with different languages and evaluation metrics, our approach outperforms all previous state-of-the-art methods.
The improvement in EM shows that our model has a stronger ability to find the correct span, due to our model making full use of the prior information of semantic structure from our coreference-ellipsis-oriented query template. 
On the Chinese datasets Table \ref{tab_3} and Table \ref{tab_4}, 
QUEEN outperforms previous methods. Since Chinese is a pro-drop language where coreference and ellipsis often happen, the improvement confirms that QUEEN is superior in finding the correct ellipsis and referring back positions.
The results on data sets of 
different domains and languages also show that our model is robust and effective.

\noindent
\textbf{Ablation Study} \quad To verify the effectiveness of the query in our proposed model and different modules, we present an ablation study in Table \ref{tab_ablation}. It is clear that query is important to improve performance on all evaluation metrics. Meanwhile, only using coreference-oriented or ellipsis-oriented template still improves the performance, as it can also bring semantic structure information.

\noindent
\textbf{Inference Speed} \quad Meanwhile, to compare the inference speed between the current fastest model RUN and our Edit Operation Scoring Network, we conduct experiments using the code released
\footnote{https://github.com/microsoft/ContextualSP/tree/master/\\incomplete\_utterance\_rewriting}. Both models are implemented in PyTorch on a single NVIDIA V100. The batch size is set to 16. Meanwhile, In order to fairly compare the speed of the two networks, we performed Distant Supervision and Query Construction before comparing.
The results are shown in Table \ref{tab_sp2}.

\begin{table}
\centering
\setlength{\tabcolsep}{2.6mm}{
\renewcommand{\arraystretch}{1.1}
\begin{tabular}{ccc}
\hline
{Model} & {Inference Speed} & {Speedup}\\
\hline
{RUN+BERT} & {1.69it/s} & {1.00$\times$}\\
{\textbf{QUEEN}} & {\textbf{2.13it/s}} & {\textbf{1.26$\times$}}\\
\hline
\end{tabular}}
\caption{{The inference speed comparison between RUN and QUEEN on REWRITE dataset.}}
\label{tab_sp2}
\end{table}

\noindent
\textbf{Case Study} \quad We also conduct case study for our proposed model.
Our model avoids the uncontrolled situations that the generation-based model is prone to, and our model can more easily capture the correct semantic span.
Table \ref{tab_6} gives 3 examples that indicate the representative situations as \citet{hao-etal-2021-rast}. The first example illustrates the cases when RUN inserts unexpected characters into the wrong places. T-Ptr-Gen just copies the incomplete utterance. Due to our generated query, the position that needs to be inserted has been explicitly promoted by the query.
The second example shows a common situation for generation-based models. T-Ptr-Gen messes up by repeating stupidly. However, this situation doesn't happen to our model, as it is not a generation-based model. 
The last example refers to a long and complex entity. For these cases, it is easier for our model to get the correct span. This is because our model learns the span boundaries from the edit operation matrix.
Compared to the generation-based model, we don't generate sentences from scratch and this reduces the difficulty. Meanwhile, our model is not based on CNN as RUN, which suffers from the limitation of receptive-field to find a longer span. 
\end{CJK}
\section{Conclusion}

\noindent
We propose a simple yet effective query-enhanced network for IUR task.
Our proposed well-designed query template explicitly brings guided semantic structural knowledge between the incomplete utterance and the rewritten utterance. 
Benefiting from extra semantic structural information from proposed query template and well-designed edit operation scoring network, QUEEN achieves state-of-the-art performance on several public datasets.
Meanwhile, the experimental results on data sets of different domains and languages also show that our model is robust and effective.
Overall, experiments show that our proposed model with this well-designed query achieves promising results than previous methods.

\section{Acknowledgements}
\noindent
This paper is supported by the National Science Foundation of China under Grant No.61876004, 61936012, the National Key R\&D Program of China under Grand No.2020AAA0106700.

\section*{Ethics Consideration}
\noindent
We use public datasets to perform our experiments.
The used open-source tools are freely accessible online without copyright conflicts.

\section*{Limitation}
\noindent
One limitation of current edit-based IUR models, is that only tokens that have appeared in the history dialogue can be selected. Therefore, these models, including ours, cannot generate novel words, e.g., conjunctions and prepositions, to cater to other metrics, like fluency. However, this can be alleviated by incorporating an additional word dictionary as \citet{see2017get} and \citet{qian2020incomplete} deals with the out-of-vocabulary (OOV) words to improve fluency. For fairness, we keep the same words during the experiment as RUN to mitigat it. We will consider this question as a promising direction for future works. 

\begin{CJK*}{UTF8}{gbsn}
\begin{table*}
\small
\centering
\setlength{\tabcolsep}{2mm}{
\renewcommand{\arraystretch}{1.3}
\begin{tabular}{cc}
\multicolumn{2}{c}{Example \# 1}\\
\hline
\multirow{2}{*}{Historical Utterance 1} & {我意见很大} \\
&{I have a lot of complaints} \\
\hline
\multirow{2}{*}{Historical Utterance 2} & {有意见保留} \\
& {Keep it yourself if there's any} \\
\hline
\multirow{2}{*}{Incomplete Utterance} & {不想保留} \\
& {Don't want to keep it myself} \\
\hline
\multirow{2}{*}{Gold} &{不想保留意见}\\
& {Don’t want to keep the complaints myself} \\
\hline
\multirow{2}{*}{T-Ptr-Gen} & {不想保留} \\
& {Don't want to keep it myself} \\
\hline
\multirow{2}{*}{RUN} & {意见不想意保留} \\
& {Complaints don't want to keep the complaints myself} \\
\hline
\multirow{2}{*}{\textbf{QUEEN}} & {不想保留意见} \\
& {Don't want to keep the complaints myself} \\
\\
\multicolumn{2}{c}{Example \# 2}\\
\hline
\multirow{2}{*}{Historical Utterance 1} & {你帮我考雅思} \\
& {Please help me on IELTS} \\
\hline
\multirow{2}{*}{Historical Utterance 2} & {雅思第一项考什么} \\
& {What is tested first for IELTS} \\
\hline
\multirow{2}{*}{Incomplete Utterance} & {考口语啊} \\
& {It's oral test} \\
\hline
\multirow{2}{*}{Gold} & {雅思第一项考口语啊} \\
& {It's oral test for IELTS} \\
\hline
\multirow{2}{*}{T-Ptr-Gen} & {考口语考口语啊} \\
& {It's oral test oral test} \\
\hline
\multirow{2}{*}{RUN} & {雅思第一项考口语啊} \\
&{It's oral test for IELTS} \\
\hline
\multirow{2}{*}{\textbf{QUEEN}} &{雅思第一项考口语啊}\\
& {It's oral test for IELTS} \\
\\
\multicolumn{2}{c}{Example \# 3}\\
\hline
\multirow{2}{*}{Historical Utterance 1} & {帮我找一下西安到商洛的顺风车} \\
& {Can you help me find a free ride from Xi'an to Shangluo} \\
\hline
\multirow{2}{*}{Historical Utterance 2} & {哪的} \\
& {Where is it} \\
\hline
\multirow{2}{*}{Incomplete Utterance} & {能不能找到} \\
& {Can you find any} \\
\hline
\multirow{2}{*}{Gold} & {能不能找到西安到商洛的顺风车} \\
& {Can you find any free ride from Xi'an to Shangluo} \\
\hline
\multirow{2}{*}{T-Ptr-Gen} & {能不能找到商洛的顺风车} \\
& {Can you find any free ride to Shangluo} \\
\hline
\multirow{2}{*}{RUN} & {能不能找到商洛的顺风车} \\
& {Can you find any free ride to Shangluo} \\
\hline
\multirow{2}{*}{\textbf{QUEEN}} &{能不能找到西安到商洛的顺风车}\\
& {Can you find any free ride from Xi'an to Shangluo} \\

\end{tabular}}
\caption{Case Study}
\label{tab_6}
\end{table*}
\end{CJK*}
\newpage


\bibliography{anthology,custom}
\bibliographystyle{acl_natbib}

\end{document}